\documentclass[10pt,twocolumn,letterpaper]{article}

\usepackage{iccv}
\usepackage{times}
\usepackage{epsfig}
\usepackage{graphicx}
\usepackage{amsmath}
\usepackage{amssymb}
\usepackage{multirow}

\usepackage[pagebackref=true,breaklinks=true,letterpaper=true,colorlinks,bookmarks=false]{hyperref}

\iccvfinalcopy 

\usepackage{mmstyle}

\graphicspath{{figures/}}

\def\iccvPaperID{991} 

\ificcvfinal\fi
\begin{document}

\title{From Trailers to Storylines: An Efficient Way to Learn from Movies}

\author{
	Qingqiu Huang, ~~ Yuanjun Xiong, ~~ Yu Xiong, ~~ Yuqi Zhang, ~~ Dahua Lin \\
	Department IE, The Chinese University of Hong Kong \\
	{\tt\small \{hq016, xy012, xy017, zy016, dhlin\}@ie.cuhk.edu.hk}
}


\makeatletter
\let\@oldmaketitle\@maketitle
\renewcommand{\@maketitle}{\@oldmaketitle
	\vspace{-10pt}
	\includegraphics[width=\linewidth]{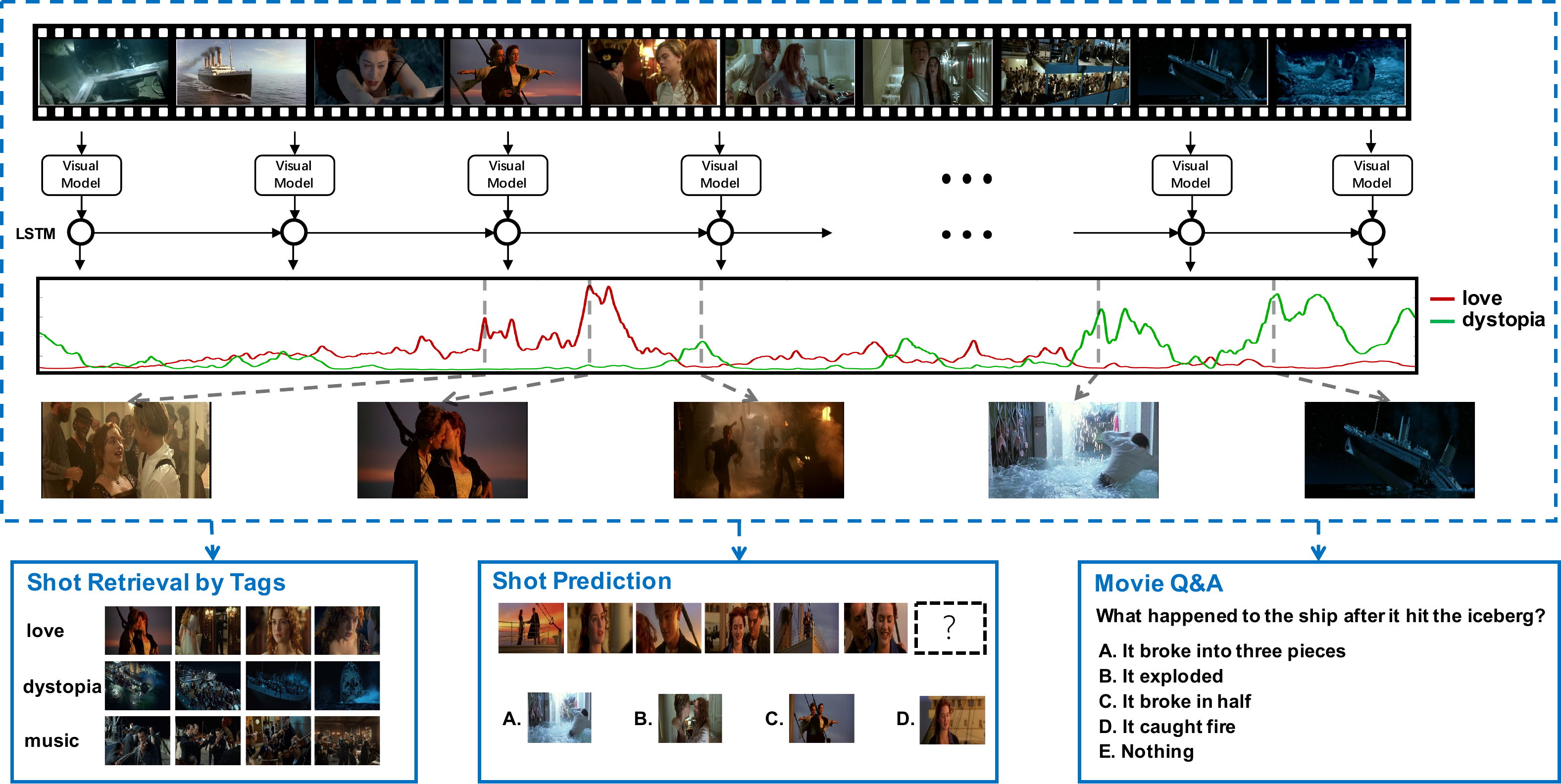}
	\small{ 
		Figure 1. We propose an effective and efficient framework to analyze movies by
		learning visial model from trailers and reconstructing temporal structure in movies,
		which can not only predict shot-level tags, although supervised by just video-level tags,
		but also apply to many tasks of movie analysis,
		like shot retrieval, shot prediction and movie Q\&A.
	}
	\bigskip}
\makeatother

\maketitle



\begin{abstract}
The millions of movies produced in the human history are valuable resources
for computer vision research.
However, learning a vision model from movie data would meet with
serious difficulties.
A major obstacle is the computational cost -- the length of a movie is
often over one hour, which is substantially longer than the short video
clips that previous study mostly focuses on.
In this paper, we explore an alternative approach to learning vision models
from movies. Specifically, we consider a framework comprised of
a visual module and a temporal analysis module.
Unlike conventional learning methods, the proposed approach learns these
modules from different sets of data -- the former from trailers while
the latter from movies.
This allows distinctive visual features to be learned within
a reasonable budget while still preserving long-term temporal
structures across an entire movie.
We construct a large-scale dataset for this study and define a
series of tasks on top. Experiments on this dataset showed that
the proposed method can substantially reduce the training time
while obtaining highly effective features and coherent temporal
structures.
\end{abstract}


\section{Introduction}
\label{sec:introduction}

{\em ``Promise me you'll survive ...
That you won't give up, no matter what happens. No matter how hopeless.''}
This heartbroken moment in \emph{Titanic},
the epic romance-disaster film directed by James Cameron,
has deeply moved everyone in front of the screen.
Movies contain tremendous values -- they are not only a form of entertainment,
but also a rich medium that reflects our culture, society, and history.
From the standpoint of computer vision research,
they also constitute a valuable source of data for visual learning,
from which we can learn how various phenomena, situations,
and even feelings can be presented in a visual way.

The value of movies has been noticed by the computer science
community since long ago.
Over the past decades, numerous studies have been
done to analyze movies from different perspectives.
\cite{bamman2014learning,weng2009rolenet,park2009character} study movie
from its characters by either building a role-net of the characters or
learning the hidden persona of the characters.
\cite{zhu2015aligning} try to align movies and books,
which is way towards story-like visual explanation.
\cite{MovieQA} set up a Q\&A benchmark consists of more than $10,000$
questions with high semantic diversity
and provides different sources of a movie, like plot,
to learn visual question answering models.
Yet, an important question regarding the movie data
has rarely been explored --
{\em can we learn a vision model for movie understanding?}

%
In this study, we aim to explore an effective approach to movie
understanding, one that goes from low-level feature representation
to high-level semantic analysis.
Towards this goal, we are facing two significant challenges, namely
the prohibitive cost in \emph{computation} and \emph{annotation}.
First, current video analysis research~\cite{WangS13IDT,
Simonyan14TwoStream,Tran15C3D,WangQT15TDD,Wang2016TSN,Donahue2015LRCN}
focuses on short video clips, \ie~those lasting for seconds
or at most several minutes.
In contrast, movies usually last for substantially longer,
\eg~one hour or more.
For videos of this scale, even simple processing, \eg~extracting CNN features,
may take an unusually long time, let alone training a model thereon.
On the other hand, vision models, \eg~convolutional networks, require
a large number of \emph{annotated} samples to train.
Obtaining annotated training samples, even for images, is widely known as
a costly procedure.
It goes without saying that it would be a formidable task to
annotate movies, which contain much more complicated structures.

%
Our approach to tackle these difficulties is inspired by
an important fact -- movies often come with trailers.
\emph{Trailers} are short previews of movies, which
often contains the most significant shots selected by professionals.
Therefore, from a diverse set of trailers, one can see a wide range of
representative shots and thus learn useful visual cues for movie analysis.
Also, trailers are much shorter than movies,
often lasting for less than five minutes.
Hence, learning models from trailers should be affordable
if done efficiently.
Whereas a \emph{trailer} preserve significant visual features of a movie,
it losts a key aspect -- the temporal structure.
Particularly, a movie often presents a story in a natural way following
a logical storyline, while a trailer is just a compilation of
distinctive shots, which can be far from each other in the original timeline.
Hence, we can not expect to learn the temporal structures from trailers.

%
With the complementary natures of movies and trailers in mind,
we explore an alternative approach to \emph{efficiently} learning from movies.
Specifically, the proposed framework integrates two key modules:
a \emph{visual analysis} module learned from \emph{trailers},
and a \emph{temporal analysis} module learned from \emph{movies}
but on top of the features extracted by the visual module.
A key observation behind this design is that state-of-the-art models
for visual extraction, \eg~convolutional networks~\cite{krizhevsky2012imagenet},
are typically much heavier than temporal models, \eg~Markov chain
or recurrent networks~\cite{hochreiter1997long}. Hence, by learning these components
from different sources of data, we can maintain the computational cost
at an affordable level while allowing the framework to capture
long-time temporal structures across a movie.
Further more, to train these models, we introduce two strategies:
(1) mining meta-data, \eg~the information about movie genres and
plot keywords, to supervise the learning of the visual module; and
(2) learning the temporal module in a \emph{self-supervised} manner,
that is, omitting parts of the chain and letting the model to
predict them given the rest.

%
To support this study and facilitate future research along this direction,
we construct \emph{LSMTD}, a large-scale movie and trailer data set, which
contains $508$ full movies and $34,219$ trailers.
The total length of these videos is more than $2,200$ hours.
Thereon we define a series of tasks to evaluate the capability of movie analysis methods,
\eg to choose the ($m+1$)-th shot from the candidates given a $m$ shots sequence.
Our experiments on this dataset showed that the proposed approach substantially
reduce the training time, while still outperform the model trained in
conventional ways on various tasks.

%
To sum up, the contributions of this work mainly lie in two aspects:
(1) We propose an alternative way to learn models for movie understanding,
where the visual module and the temporal analysis module are respectively
trained on trailers and movies, using meta-data and self-supervised learning.
(2) We construct a large-scale movie and trailer dataset, define a series
of tasks to assess the capability in movie understanding, and thereon
perform a systematic study to compare different training strategies.


\section{Related Works}
\label{sec:related}

\paragraph{Studies on Movies.}
Due to the rich content and meta information, movies have been a gold mine for AI research.
Numerous attempts have been make to automatically understand movies from different aspects.
A stream of work focuses on the movie characters~\cite{bamman2014learning,weng2009rolenet,park2009character},
trying to understand the relationships among them based on textual materials.
Zhu \etal~\cite{zhu2015aligning} proposed a method to align movies and books,
in order to learn high-level semantics therefrom.
Tapaswi \etal~\cite{MovieQA} developed a Q\&A benchmark, which suggests an alternative
way to learn from movies, that is, via visual question answering.
These works mostly rely on text-based information, \eg~plot, subtitles, and scripts,
to mine and learn the semantics.
When they try to incorporate visual observations,
they simply use the features extracted using a convolutional network trained on
image-based tasks (rather than on the movies).
This is largely due to the computational difficulties of training on the movies themselves.

\vspace{-10pt}
\paragraph{Studies on Trailers.}
There have also been studies done on trailers.
These studies often focus on other problems, \eg~face recognition~\cite{Ortiz2013Face}.
There are also efforts trying to generate trailers for user-uploaded videos by learning from structures of movies~\cite{Kang2006Montage,Oos2016Semantic}.
In \cite{simoes2016movie,zhou2010movie}, the genres classification problem is proposed to be tackled with trailers. For this purpose, datasets with several thousand trailers have been constructed.
These works are all based on the trailers themselves without considering their corresponding movies.
It is noteworthy that movies and trailers have rarely been considered together in vision research.
To our best knowledge, this work is the first practical approach that bridges trailers and movies
and allows knowledges learned from trailers to be transferred to full movie analysis.

\vspace{-10pt}
\paragraph{Video-based Recognition.}
Video-based recognition is also an active area in computer vision.
Related topics include action recognition and video summarization.
Over the past several years, the action recognition task has witnessed remarkable
progress~\cite{WangS13IDT,Simonyan14TwoStream,Tran15C3D,WangQT15TDD,Wang2016TSN,Donahue2015LRCN},
thanks to the advances in deep learning.
For this task, following earlier efforts presented in \cite{Simonyan14TwoStream} and~\cite{WangQT15TDD},
recent studies have gradually shifted from
hand-crafted features~\cite{WangS13IDT} to representations learned with deep convolutional networks.
A key issue in video analysis is the excessive computational cost.
Recently, Temporal Segment Networks (TSN)~\cite{Wang2016TSN} was proposed, which tried
to tackle this problem with sparsely sampled frames.
Another important family of work is video summarization,
which aims to find the most representative frames or snippets to represent a video.
This problem is usually solved using unsupervised or weakly supervised learning~\cite{Potapov2014Category}.
A comprehensive review of this topic is provided in~\cite{Truong2007Video}.
Whereas the proposed model provides the functionality of identifying representative frames
as a byproduct,
our ultimate goal is different, that is, to learn powerful visual representations and
the temporal structures of movies, instead of just choosing representative frames.


\section{Large-Scale Movie and Trailer Dataset}
\label{sec:dataset}

\begin{figure}[t]
	\centering
	\setcounter{figure}{1}
	\includegraphics[width=\linewidth]{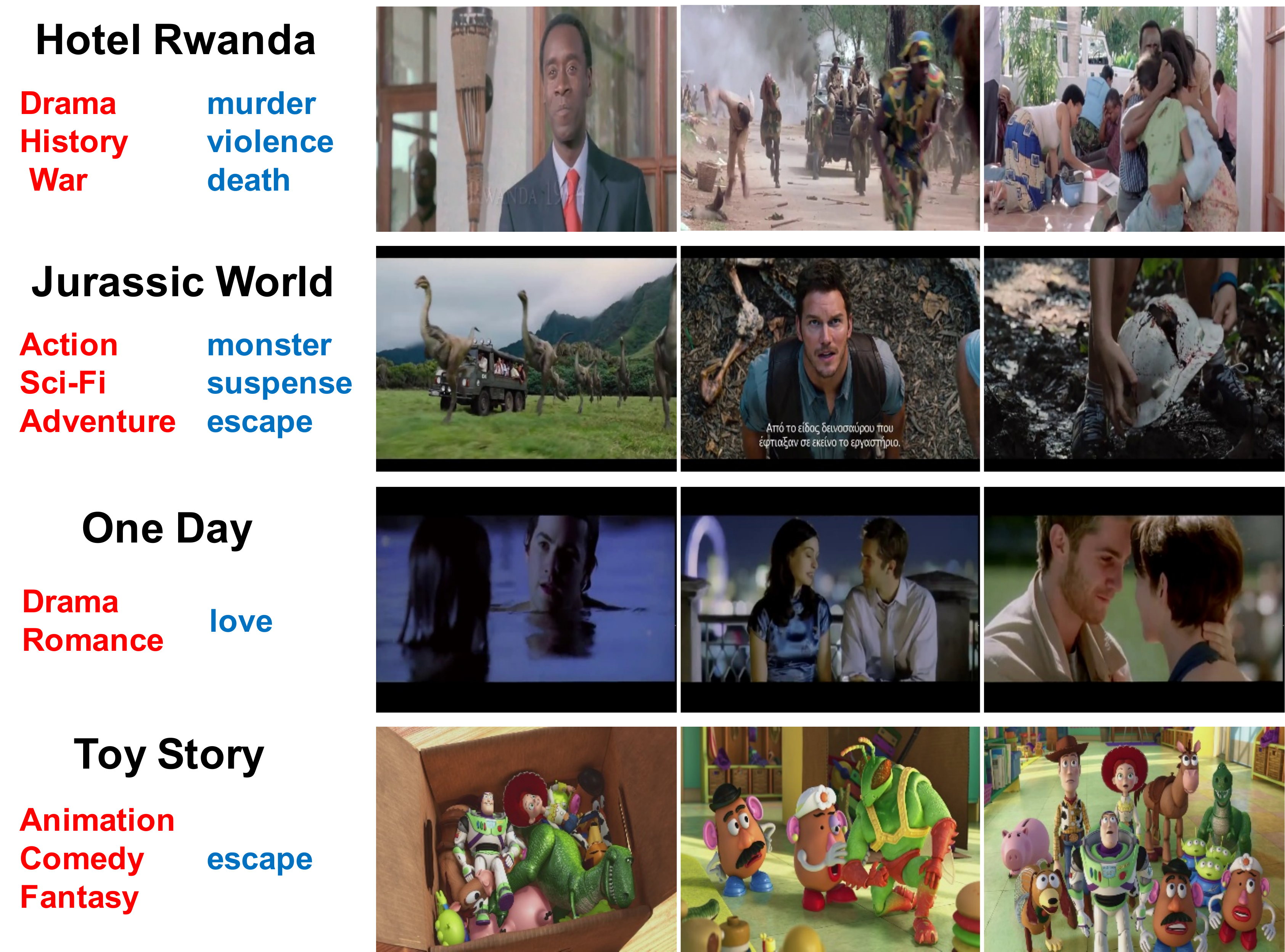}
	\caption{\small
		Samples of LSMTD.
		The black words are the move titles,
		the red ones are their genres, and
		the blue ones are plot keywords.
	}
	\label{fig:dataset}
\end{figure}

\begin{table}[t]
	\centering
	\caption{The basic statistics of several trailer dataset. }
	\vspace{0.1cm}
	\begin{tabular}{|c|c|c|c|}
		\hline
		Name & MGCD\cite{zhou2010movie} & LMTD\cite{simoes2016movie} & LSMTD \\ \hline
		trailers &  1239 & 3500 & 34,219 \\ \hline
		movies & - & - & 508  \\ \hline
		genres & 4 & 22 & 22  \\ \hline
		plot keywords & - & - & 33  \\ \hline
		duration(h) & 42 & 118 & 2,200 \\ \hline
	\end{tabular}
	\label{tab:dataset}
\end{table}

This work aims to learn models from \emph{both} movies and trailers.
For this purpose, existing datasets~\cite{zhou2010movie,simoes2016movie}
are limited. They contain several thousands of trailers along with
some meta information, \eg~genres, but full movies were not provided.
To facilitate our study, we constructed a new dataset, named
\emph{Large-Scale Movie and Trailer Dataset (LSMTD)}.
This dataset contains $34,219$ movie trailers
(around $1$ billion frames, $200$ million shots)
and $508$ movies (around $1$ billion frames, $100$ million shots).
The movies and trailers in LSMTD together contain over $2200$ hours
of video materials and more than $2$ billion frames.
To our best knowledge, this is the largest dataset that has ever been built
for visual analysis of movies,
and one of the largest if taking all video datasets into account.



We purchase the movie DVDs from Amazon and other commercial channels,
download the trailers from YouTube, collect their genres
from IMDB\footnote{IMDB: \url{http://www.imdb.com}}, and
user-provided plot keywords from TMDB\footnote{TMDB: \url{https://www.themoviedb.org}}.
Every trailer and movie in LSMTD is associated with at least one genre.
Contrary to genres, \eg~\emph{``Action''} and \emph{``Drama''},
plot keywords, \eg~\emph{``love''}, \emph{``escape''}, and \emph{``alien''},
are often more specific in describing the movie content.
The plot keywords are very sparse --
there are more than $20K$ distinct keywords for $34K$ movies,
most of them appear for just one or two times.
To obtain a meaningful set of keywords,
we filter out infrequent keywords and merge synonyms and closely related concepts,
such as \emph{``blood''} and \emph{``gore''}, which results in a diverse set of $33$
unique keywords.

Among all trailers and movies,
$207$ movies and $5,601$ trailers are associated with at least one keyword in this set.
Table~\ref{tab:dataset} shows some statistics of the LSMTD in comparison with
other datasets.
Figure~\ref{fig:dataset} shows several examples of the trailers together with their
genres and plot keywords.

We plan incorporate extra meta-information into the dataset and
will release the dataset to promote further study
of movie analysis. Due to legal constraints, trailers and movies
will be released in the form of urls to the sources.


\section{Movie Analysis Framework}
\label{sec:framework}

We aims to learn both visual information and temporal structures from the movies.
What's more, we want to learn it in a very \emph{efficient} way.
This means we do not expect to take all frames of an entire movie in one step of learning,
which is both prohibitively expensive (due to the sheer volume of data contains in a movie)
and unnecessary (frames in a movie are highly redundant).
Instead, we take \emph{shots} as the units, motivated by the observation that
frames within a shot are highly similar.

Based on shots, the proposed framework decompose the task of learning into two parts,
that is, to learn visual representations from trailers and temporal structures
from movies. The rationales behind this design mainly consist in two aspects.
On one hand, trailers are much shorter than the movies
while usually containing the most significant shots of the movies.
This condensed collection of distinctive shots forms a good basis
where we can learn reasonable representation from within.
On the other hand, a trailer is usually formed by shots sparsely
selected from the a movie.
Therefore, it does not preserve the temporal structures of the original storyline,
which is also an important aspect in movie understanding.
To capture the temporal structures, we formulate an LSTM-based model on top of the visual features,
and learn it in a \emph{self-supervised} way.
In what follows, we will describe these two components in turn.

\subsection{Learning Visual Representations from Trailers}
\label{subsec:framework_trailer}

\begin{figure}[t]
	\centering
	\includegraphics[width=\linewidth]{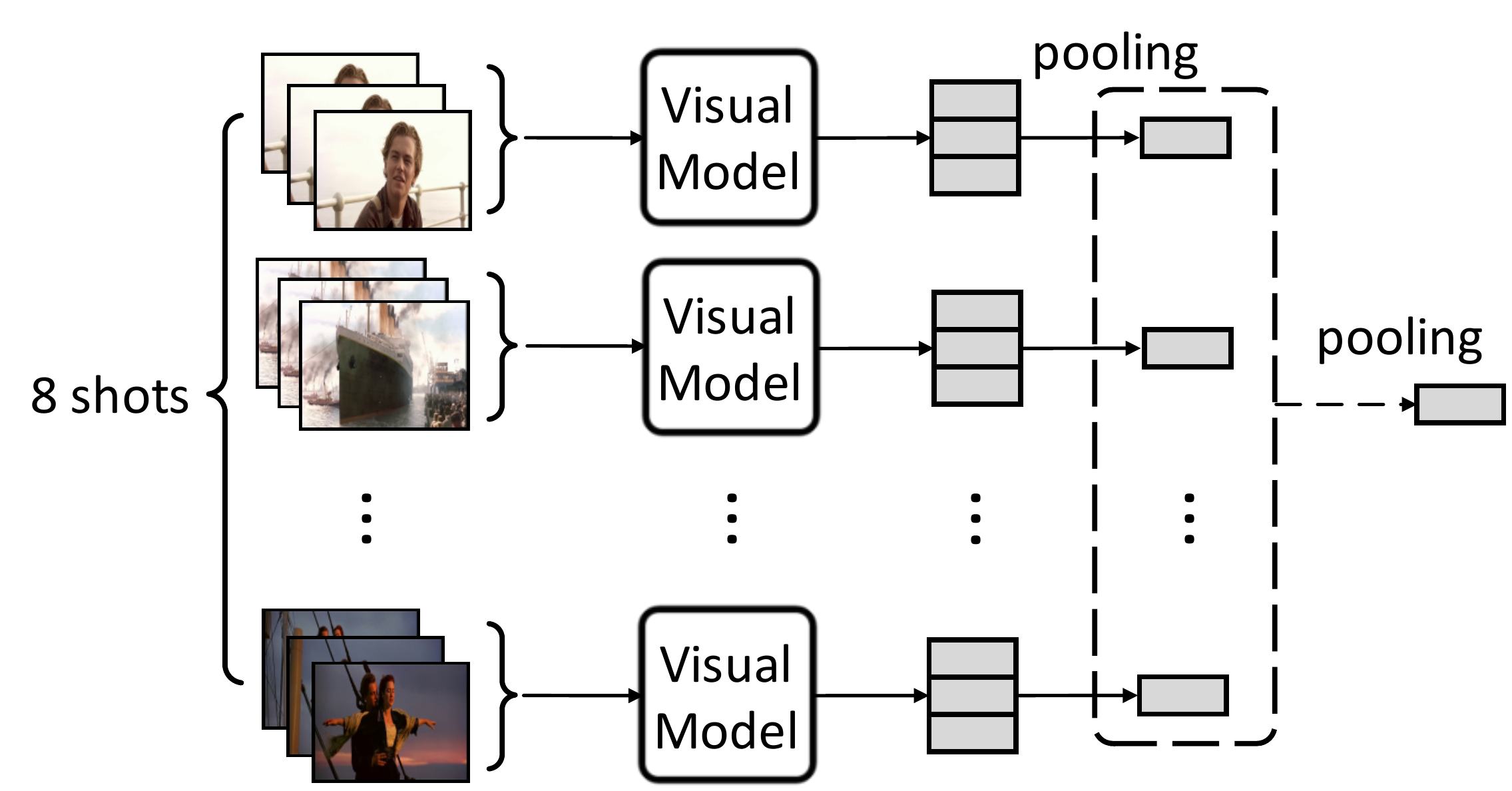}
	\caption{\small
		Learning Visual Representations from Trailers.
	}
	\label{fig:framework_trailer}
\end{figure}

\begin{figure*}[t]
	\centering
	\includegraphics[width=\linewidth]{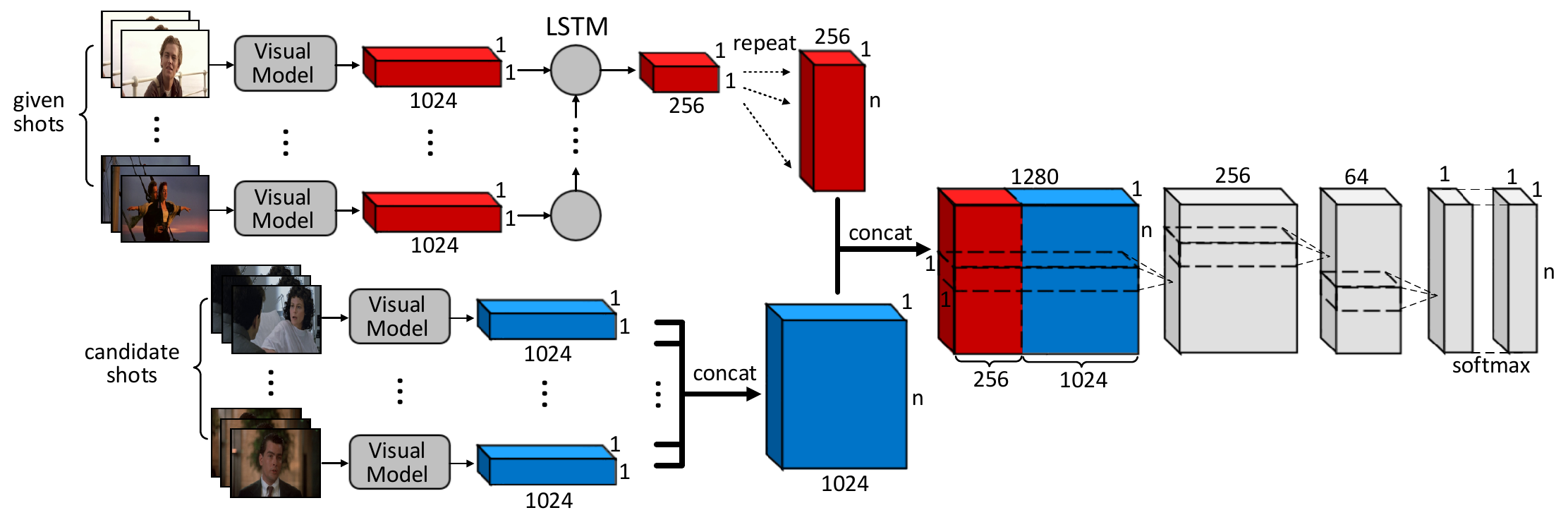}
	\caption{\small
		Learning Temporal Structures from Movies.
	}
	\label{fig:framework_movie}
\end{figure*}

We start with learning a visual model from trailers.
Specifically, the model is expected to output visual features to represent shots of the movies.
Therefore it can be seen as a \emph{shot encoder}.
Following previous work in video analysis~\cite{Wang2016TSN},
we formulate the model as shown in Figure~\ref{fig:framework_trailer},
which encodes a shot in two steps: deriving frame-wise features
via a convolutional network, and then combining them into
a shot-based representation.
The key question here is how to learn this model effectively
and efficiently.

\vspace{-10pt}
\paragraph{Shot-based Representations.}
While trailers are generally much shorter than a full movie,
\eg~$3$ minutes vs $90$ minutes, they are still substantially longer
than the video clips used in much of the previous work on video analysis,
which are usually less than $10$ seconds.
Therefore, on trailers, frame-by-frame analysis is still excessively expensive.
However, unlike ordinary videos such as those from surveillance systems,
a trailer usually comprises a sequence of distinctive shots, each lasting for
a few seconds. The frames in a shot often look quite similar.
This special property in temporal structure
suggests an efficient way to analyze trailers, that is,
to consider a trailer as a sequence of shots and sample frames sparsely
in each shot.

Specifically, given a trailer, we first use an external tool
to partition it into a sequence of coherent shots~\cite{sidiropoulos2011temporal}:
$S_1, \ldots, S_N$.
From each shot $S_i$, we sample $m$ frames: $I_{i1}, \ldots, I_{im}$.
and use a feature extractor to derive features for the sampled frames,
denoted as $\vf(I_{ij}; \vtheta)$, where $\vtheta$ is the parameter
of the feature extractor.
Then, we can derive a shot-based representation by average pooling:
\begin{equation}
	\vf(S_i; \vtheta) = \frac{1}{m} \sum_{j=1}^m \vf(I_{ij}; \vtheta).
\end{equation}
Considering the strong expressive power of convolutional neural networks (CNNs)
in visual representation, we suggest using a CNN for feature extraction.
Particularly, we use BN-Inception~\cite{Ioffe2015BN} in our experiments,
which we empirically found as a good balance between performance and cost.
Also, we set $m = 3$, which allows more robust representation as opposed
to a single frame. We also found that increasing $m$ to a greater value is unnecessary,
due to the high redundancy among frames.

\vspace{-10pt}
\paragraph{Training Visual Models}
We rely on associated tags, \eg~genres and plot keywords,
to supervise the learning of the visual model.
These tags are attached to the entire movie/trailer
instead of individual shots.
Hence, to learn from such tags,
we need first combine the shot-based representations to generate
video-level predictions.

It is noteworthy that movie producers usually choose the most
representative shots for a trailer. Therefore, the shots
in the trailer are probably relevant to the genres/keywords.
This observation suggests a simple scheme to learn
the model.
Specifically, we can just sample $n$ shots from a trailer
and aggregate their features via another level of average pooling:
\begin{equation}
\vf(T; \vtheta) = \frac{1}{n} \sum_{j=1}^n \vf(S_{\phi(j)}; \vtheta).
\end{equation}
Here, $T$ denotes the trailer, and $\phi(1), \ldots, \phi(n)$ denote
the indexes of the sampled shots, and
we chose $n = 8$ to fit the training process in a GPU.
Then, via a fully connected layer and a softmax layer, we can turn
$\vf(T; \vtheta)$ into categorical predictions over the tags, and
consequently the model can be trained based on the groundtruth tags.

\textbf{Discussion:}
Here, we do not assume that each of the sampled shot is pertinent
to the given tags. As long as a tag is relevant to a few of the
samples, it will get a high prediction score -- this is very likely
due to the nature of trailers. In a certain sense, the process of
choosing shots to form a trailer, which is done by the movie producer,
can be considered as a weak form of supervision, and we are leveraging
it for free. In practice, we found that the scheme described above,
while simple, is very effective.

\begin{table*}[t]
	\centering
	\caption{Tag Classification Results. }
	\vspace{0.1cm}	
	\begin{tabular}{|c|cc|cc|cc|cc|}
		\hline
		\multirow{3}{*}{Train Data} & \multicolumn{4} {c|} {Score Average} & \multicolumn{4} {c|} {Feature + LSTM} \\
		\cline{2-9}
		& \multicolumn{2} {c|} {Genres} & \multicolumn{2} {c|} {Keywords} & \multicolumn{2} {c|} {Genres} & \multicolumn{2} {c|} {Keywords} \\
		\cline{2-9}
		& recall@3  & MAP & recall@3  & MAP & recall@3  & MAP & recall@3  & MAP \\
		\hline
		Image-base  & -				 & -			  & -		       & -				& 0.477 		 & 0.472		  & 0.199 		   & 0.127 \\
		Movie 361   & 0.433          & 0.342          & 0.192          & 0.103 			& 0.432 		 & 0.440 	      & 0.181 		   & 0.107 \\ 
		Trailer 361 & 0.421          & 0.430          & 0.154 		   & 0.126 			& 0.435 	 	 & 0.446 		  & 0.196 		   & 0.123 \\ 
		Trailer 2K  & 0.559          & 0.538          & 0.222 		   & 0.128 			& 0.491 		 & 0.513 		  & 0.217 	  	   & 0.128 \\ 
		Trailer 10K  & \textbf{0.586} & 0.587          & 0.245 		   & 0.131 			& \textbf{0.531} & 0.523		  & 0.228 		   & 0.113 \\ 
		Trailer 33K  & 0.582          & \textbf{0.596} & \textbf{0.248} & \textbf{0.139} & 0.528 		 & \textbf{0.538} & \textbf{0.236} & \textbf{0.139} \\ \hline 
	\end{tabular}
	\label{tab:exp_tag}	
\end{table*}

\begin{table}[t]
	\centering
	\caption{The duration and shots of different training sets. }
	\vspace{0.1cm}	
	\begin{tabular}{|c|cc|}
		\hline
		Name & Duration(h) & shots(k) \\ \hline
		Movie 361 &  770 & 657  \\ 
		Trailer 361 & 13 & 21 \\ 
		Trailer 2K & 70 & 117 \\ 
		Trailer 10K & 370 & 587  \\ 
		Trailer 33K & 1121 & 1761 \\ \hline
	\end{tabular}
	\label{tab:training_set}	
\end{table}

\subsection{Learning Temporal Structures from Movies}
\label{subsec:framework_movie}

In addition to the visual representations,
temporal structure is also a very important aspect of a movie.
As discussed before, this structure is mostly lost in the trailers.
Therefore, to learn this structure, we still need to rely on the movies themselves.
Fortunately, we do not need to start from scratch --
the visual models learned from trailers have already provided a powerful encoder for visual information.
Hence, we can model the temporal structures on top of these visual representations.

Our temporal structure model is based on \emph{Long Short Term Memory (LSTM)}~\cite{hochreiter1997long},
which has been shown to be a very effective model of sequential structures.
In particular, it is able to preserve long-range dependencies while being sensitive
to short-term changes, thus it very suited for capturing the temporal structures of movies.
This LSTM formulation takes a sequence of movie shots encoded by our visual model as input,
while trying to capture the semantics via the latent states.

To learn the LSTM model, we propose a \emph{self-supervised learning scheme}, which is
motivated by a question: {\em how can know whether a model really captures the sequential structure?}
A natural way to is to ask the model to predict what happens next conditioned on
the preceding observations.
In our context, this can be realized by requesting the model to predict the
$(m+1)$-th shot given the preceding ones. We call this task \emph{next shot prediction}.
However, synthesizing the frames of the next shot is a very difficult problem
and it is not directly relevant to our goal -- we are interested in the high-level semantics.
We circumvent this issue by reformulating the task \emph{next shot prediction}
as a multi-choice Q\&A problem.
The task now becomes
\emph{``given a sequence of $m$ shots, choose the best $(m+1)$-th shot from a pool of candidates''}.
The pool of candidates should comprise both the correct answer and a series of distracting options.
The distracting options can be chosen from either the same movie or other movies,
which can influence the difficulty of the task --
the former is obviously more difficult, as the options from the same movies tend to be more deceiving.


Generally, a multi-choice Q\&A problem can be formulated as a three-way scoring function~\cite{MovieQA},
which we denote by $s(q, a \mid C)$,
where, $q$ denotes the question, $a$ denotes a candidate answer, and
$C$ is condition, which includes other observations besides the questions and the answers.
Then, question answering can be performed by finding the answer with the maximum $s$ score
among all provided choices, as:
\begin{equation} \label{eq:qa}
	\hat{i} =\argmax_{i \in \{1 \ldots n\}} s(q, a_i \mid C).
\end{equation}
For our specific problem of ``shot prediction'',
the condition $C$ is the sequence of $m$ preceding shots,
$a$ is one of the candidate shots.
%

We formulate the scoring function $s_q$ as an LSTM constructed on discriminative CNNs,
as shown in Figure~\ref{fig:framework_movie}.
The overall pipeline can be briefly described as follows.
On top of the shot-based features derived from our visual model, we construct an LSTM,
where each time step corresponds to a shot in the movie sequence.
Particularly, at each step, an LSTM unit takes both the preceding states and the visual feature
of the current shot as input, and yield an $256$-dimensional feature, denoted as $\vu$, as an output,
which encodes the model's understanding over all the shots that it has seen.
On the other hand, each candidate shot is characterized by its own visual feature of dimension $1024$.

We repeat $\vu$ by $n$ times and concatenate them with the candidate shot representations,
thus forming a matrix of size $n \times 1280$, where each row correspond to a candidate combined
with the condition $\vu$. This combined representation, through a series of $1 \times 1$ convolution,
will be distilled into a score vector of length $n$.
Finally, via a softmax layer, these scores will be converted into normalized probability values,
one for each choice.
We can then learn the network by maximizing the log-probabilities of the correct answers.

\textbf{Discussion:}
This framework has two noteworthy aspects.
First, the LSTM framework is based on the shot-based representations learned from the trailers.
These features can be extracted at the beginning and cached, and we won't update the underlying
CNN during training.
Therefore, it can be trained very efficiently even on entire movies.
Second, via the \emph{next-short prediction} problem, the training can be \emph{self-supervised},
without the need of any external annotations to provide supervisory signals.


\section{Experiment Result}
\label{sec:experiment}

The primary goal of this work is to develop a method
that can learn visual representations and temporal structures from both
trailers and movies.
To test the capability of the learned models, we set up two
benchmark tasks, namely
\emph{tag classification} and \emph{shot prediction},
and testing our framework on Movie QA benchmark\cite{MovieQA}
in order to evaluate the models from multiple perspectives.

We proposed to learn the visual model from trailers instead
of directly from the movies. To validate the effectiveness of this approach,
we compare different configurations, where the models are respectively learned
from movies and different subsets of the trailers.
All these models adopt the same
BN-inception architecture~\cite{Ioffe2015BN}.

In our experiments, the $508$ movies from LSMTD  are randomly divided into three sets,
$361$ for training (named ``Movie $361$''),
$41$ for validation (named ``Movie $41$'')
and $106$ for testing (named ``Movie $106$'').
As for the trailers, we first filter out the trailers whose movies are in the validation set and testing set.
The remaining ones form the set ``Trailer 33K'', which contains around $33,000$ trailers.
From this whole training set of trailers, we randomly sample $2,000$ and $10,000$ trailers
and construct two subsets ``Trailer 2K'' and ``Trailer 10K''.
We also construct a special subset ``Trailer 361'', which contains the trailers corresponding to
the $361$ training movies.
In conclusion, we get five different set of data for training, as summarized in Tab.~\ref{tab:training_set}.
These subsets are constructed to investigate the effect of number of trailers on the quality of visual models,
and compare the effectiveness of trailer-based and movie-based training.

\subsection{Tag Classification}
\label{subsec:tagging}

The first task is tag classification.
For each movie, we ask the models to predict both its genres and plot keywords.
We construct our visual models as described in \ref{subsec:framework_trailer}
and train it with both genres and plot keywords under a multi-task setting.
We train the visual models on the $5$ training sets described above,
obtaining $5$ different models.
During training, we randomly sample $8$ shots of a video (movie or trailer), and extract $3$ frames of each shot.
The output is obtained by averaging the response of all $24$ frames.
During testing, we try two different ways to predict the tags:
\textbf{(1)} \emph{Score Average}:
average all the predictions from all sampled shots to get the final prediction.
\textbf{(2)} \emph{Feature+LSTM}:
construct an LSTM on top of shot-based representations, and train the LSTM
to encode the shot sequence. The outputs of all time steps of the LSTM
are averaged to obtain the final prediction.
For the Feature+LSTM approach, in addition to the five models
trained on video subsets, we also consider a setting that uses the features extracted
directly from a CNN trained on ImageNet~\cite{ILSVRC15} (without finetuning on LSMTD).
This setting is referred to as ``Image-base''.

Tab.~\ref{tab:exp_tag} summarizes the results.
We can see that the visual model trained on the ``Trailer 33K'' performs the best,
on both genres and plot keyword predictions.
This demonstrates that as the number of trailers increase, the quality of learned visual model would also improve.
%
Interestingly, models learned on ``Movie 361'' get comparable or slightly worse results
compared with those on ``Trailer 361''.
When compared with the models trained on more trailers (the overall cost is still not as high),
the performance of ``Movie 361'' falls way behind.
This observation clearly suggests that in terms of learning visual representations,
training on trailers, which often contain the most distinctive shots,
is more effective than training directly on movies.

\subsection{Movie Q\&A}
\label{subsec:MQA}

\begin{figure}[t]
	\centering
	\includegraphics[width=\linewidth]{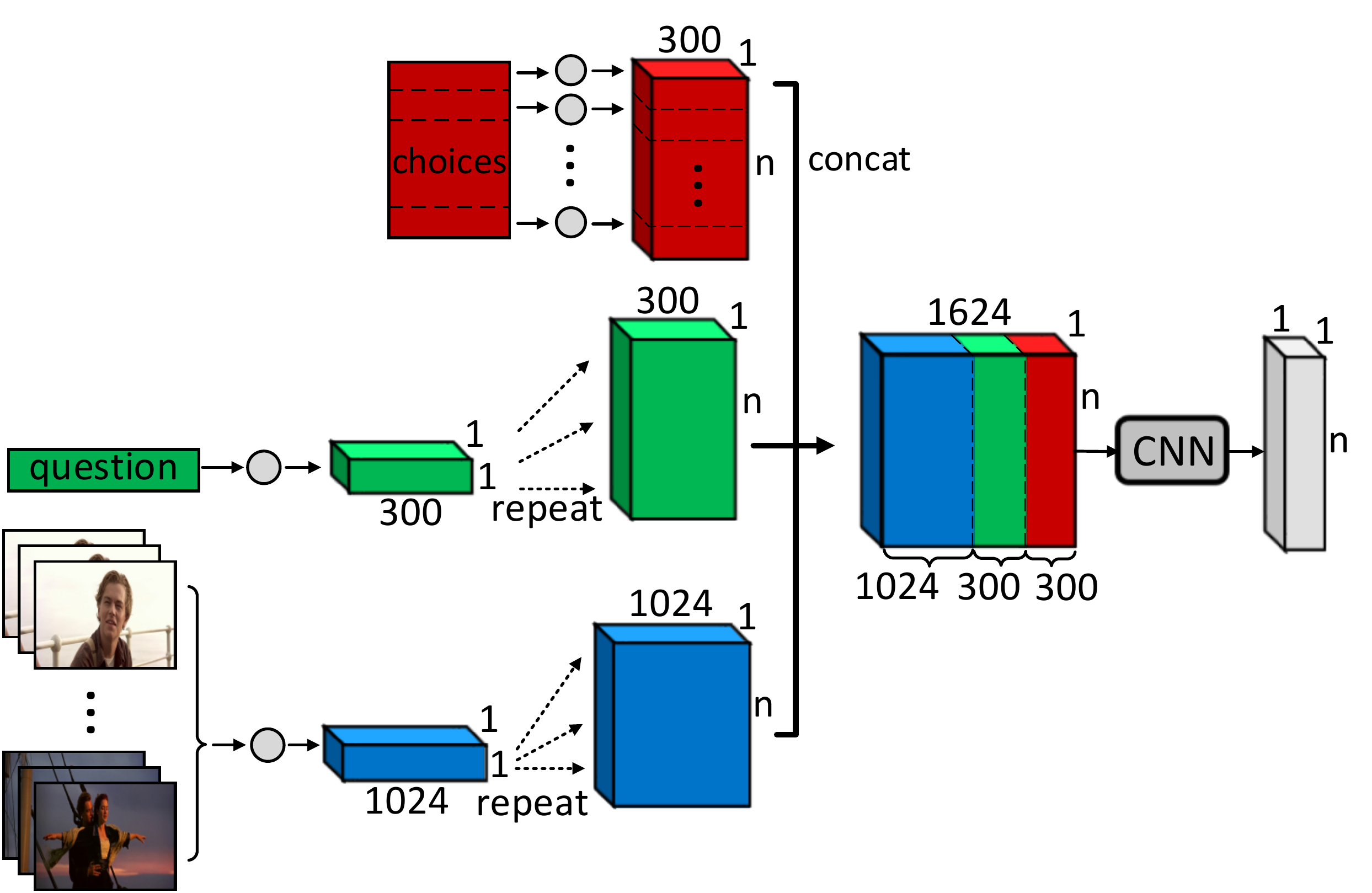}
	\caption{\small
		This figure shows the overall structure of our Q\&A model.
		At first, the features of the video clips ($1024$ dimension) and
		the embedding of the question ($300$ dimension) are concatenated.
		Then we replicate the concatenated vector for $n$ times and combine
		them with the embeddings of the choices (answer candidates) of
		dimension $300$,
		which would result in a matrix of size $n \times 1624$.
		Then we feed it to a CNN with three $1 \times 1$ convolutional layers
		followed by a softmax layer, and finally obtain a probability vector
		over the $n$ given choices.
	}
	\label{fig:qa_model}
\end{figure}

We also tested the visual model on the MovieQA benchmark dataset~\cite{MovieQA},
to evaluate how relevant the learned visual representations are to semantic understanding.
The MovieQA dataset consists of $14,944$ questions about over $400$ movies,
with one correct answer from $ 5 $ choices each.
It provides multiple sources of information, like movie clips, plots, subtitles and etc,
but not all the $294$ movies contain video clips.
As our goal here is to evaluate the visual representations,
we train and test on the subset containing video clips,
which consists of $6,462$ questions, $4,318$ for training, $886$ for validation and $1,258$ for testing.

This Q\&A problem is the same as what we formulated in~Eq.\eqref{eq:qa},
where $q$ and $a$ are the embeddings of the question and the answers,
and $C$ is the embedding of the video clip aligned with the question.
We use \emph{word2vec}~\cite{mikolov2013efficient} to embed $q$ and $a$
and the feature extracted by our visual model as the embedding of the video clips.
The overall pipeline of our Q\&A model are shown in Figure~\ref{fig:qa_model}.

Tab.~\ref{tab:exp_mqa} shows the testing results from ``Movie 361'' and ``Trailer 33K''
and compare them with SSCB, the baseline provided in~\cite{MovieQA}.
It can be seen that the features from the models learned on trailers
outperform the one on movies.
This again corroborates our hypothesis -- training on trailers is effective
for learning visual representations for movies.

\begin{table}[t]
	\centering
	\caption{Movie QA Results.}
	\vspace{0.1cm}
	\begin{tabular}{|c|c|}
		\hline
		Model & Accuracy \\ \hline
		SSCB~\cite{MovieQA} & 22.18 \\ \hline
		Movie 361 & 23.45 \\
		Trailer 33K &  24.32 \\ \hline
	\end{tabular}
	\label{tab:exp_mqa}
\end{table}

\subsection{Shot Prediction}
\label{subsec:retrival}

\begin{table}[t]
	\centering
	\caption{Shot Prediction Accuracy. }
	\vspace{0.1cm}
	\begin{tabular}{|c|cc|cc|}
		\hline
		\multirow{2}{*}{Model} & \multicolumn{2} {c|} {In Movie} & \multicolumn{2} {c|} {Cross Movie} \\
		\cline{2-5}
		& Average  & LSTM  & Average & LSTM \\
		\hline
		Movie 361   & 0.185 		 & 0.364 		  & 0.713 	 	   & 0.795 \\
		Trailer 361 & 0.181 		 & 0.362 		  & 0.637 		   & 0.792 \\
		Trailer 2K  & 0.172 		 & 0.367 		  & 0.618 		   & 0.796 \\
		Trailer 10K  & 0.161 		 & 0.399 		  & 0.551 		   & 0.817 \\
		Trailer 33K  & 0.154			 & \textbf{0.401} & 0.529		   & \textbf{0.825} \\ \hline
	\end{tabular}
	\label{tab:exp_retrival}
\end{table}

\begin{figure*}[t]
	\centering
	\includegraphics[width=0.9\linewidth]{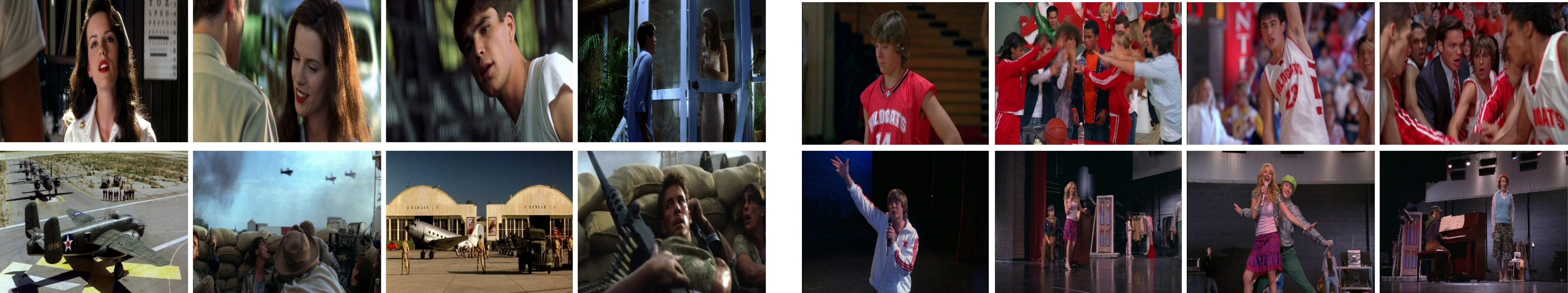}
	\caption{
		Shot Retrieval.
		The left part are the shots with the high responses
		to \emph{``Romance''}(first row) and \emph{``War''}(second row) in \emph{Pearl Harbor}.
		The right part the shots with high response
		to \emph{``Sport''}(first row) and \emph{``Music''}(second row) in \emph{High School Musical}.
	}
	\label{fig:visual_1}
\end{figure*}

\begin{figure*}[t]
	\centering
	\includegraphics[width=\linewidth]{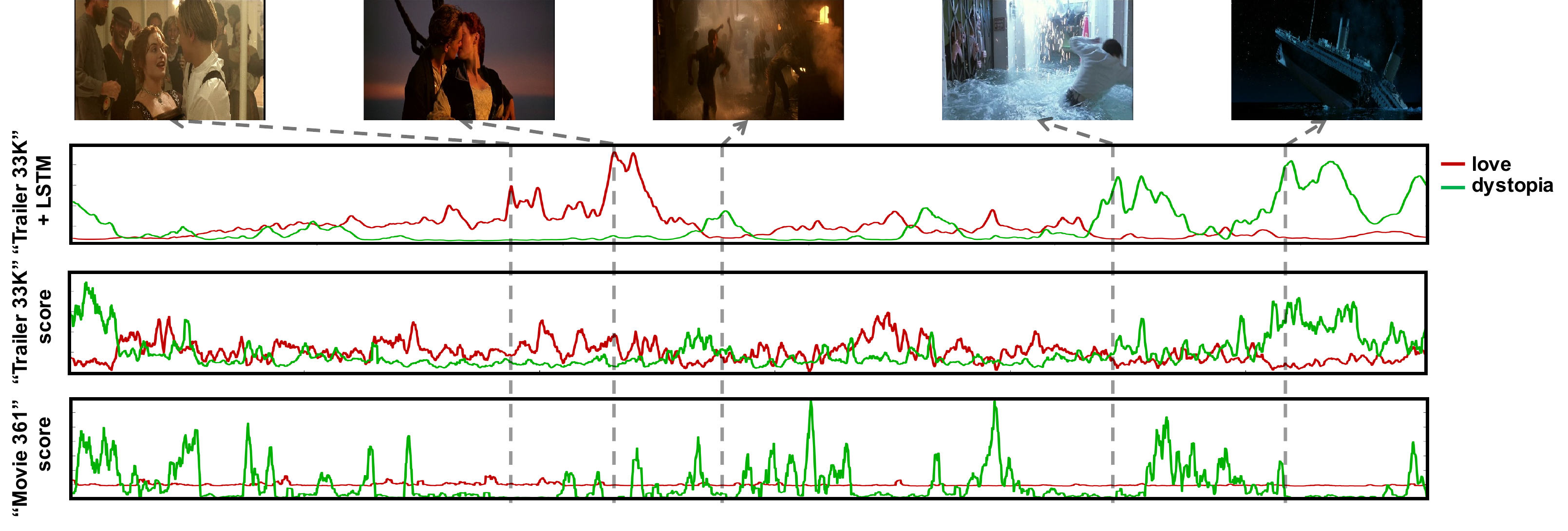}
	\caption{
		Response to the plot keywords in \emph{Titanic}.
		The three plots are  the response from visual model trained on``trailer 33K'' followed by a LSTM,
		visual model trained on ``trailer 33K''
		and visual model trained on ``movie 361''.
		From the comparison of these plots,
		we can see that learning visual information from trailers do better than from movies
		and we can reconstruct the temporal structure by a self-supervised LSTM.
	}
	\label{fig:visual_2}
\end{figure*}

Based on LSMTD, we define a shot prediction task to evaluate the capability of the temporal structure model in analyzing movies.
Given a sequence of $m$ shots, we expect the temporal model to predict the $(m+1)$-th shot.
As described in~\ref{subsec:framework_movie},
we design this task as a multi-choice Q\&A problem.
The question is ``which is the $(m+1)$-th shot given a sequence of $m$ shots'' .
The distractive choices are shots randomly picked from either the same movie or other movies.

Based on this task, we construct a benchmark on shot prediction, upon the $508$ movies of LSMTD.
This benchmark has two settings:
(1) ``In Movie'', where the distracting answers are other shots within the same movie, and
(2) ``Cross Movie'', where the distracting answers are sampled from all $508$ movies.
Each setting has $835,164$ questions,
and each question comes with a correct answer and $31$ deceiving ones.
These questions are divided into three disjoint sets:
(1) \emph{training}: $643,432$ questions from ``Movie $361$'',
(2) \emph{validation}: $74,059$ questions from ``Movie $41$'', and
(3) \emph{testing}: $117,673$ questions from ``Movie $106$''.

We compare our temporal model described in~\ref{subsec:framework_movie}
with a straightforward baseline.
This baseline simply averages the features of the given shots and calculates the cosine distances between the averaged feature vector and those from the query shots.
The results are summarized in Tab.~\ref{tab:exp_retrival}.
We can see that our temporal model significantly outperforms the baseline.
This shows that our model, in a certain way, does capture the temporal structures, which
give it the improved ability to make predictions on next shots.
We also notice that higher accuracies on the ``Cross Movie'' setting as compared to the
``In Movie'' setting. This is because the shots in the same movie are usually more confusing than those from other movies as they share similar visual styles and movie characters, etc.
Again, we still observe that the visual models learned from trailers perform better than the models trained on movies, and that the performance increases as the numbers of training trailers grows. These observations are in accordance with our earlier observations.

\subsection{Visualization}
\label{subsec:visual}


Finally, we evaluate whether the visual models
capture the connections between semantic concepts and
visual observations via a qualitative study.
In this study, we perform shot retrieval.
We select some movies and try to find the shots with the highest responses to the tags.
Figure~\ref{fig:visual_1} show some representative results.
We can see that the shots with high responses to the query tags are indeed highly relevant,
which shows that the visual models have effectively learned the semantics behind
the visual observations.

For comparison, Figure~\ref{fig:visual_2} shows the responses of the shots in \emph{Titanic}
to two plot keywords, \emph{``love''} and \emph{``dystopia''}, based on three models respectively.
We can observe accurate responses produced by the model trained on \emph{``Trailer 33K''},
while the one trained on \emph{``Movie 361''} failed to capture these semantic concepts,
and producing poorly aligned responses.
More results will be provided in the supplemental materials.


\section{Conclusions}
\label{sec:conclusion}


This paper presented an efficient approach to learn visual models from movies.
Particularly, it learns the visual representations from trailers, taking advantage
of the trailers' distinctive natures, and learns the temporal structures
from movies via a self-supervised formulation.
We collected a large scale movie and trailer dataset, which contains over $34K$ trailers
and $508$ movies, and defined two tasks thereon, namely tagging and shot prediction,
to evaluate a model's capability in understanding a given movie purely from visual observations.
We also test our framework on the movie Q\&A task.
Experimental results on all three tasks consistently showed the effectiveness of the proposed framework.

{\small
\bibliographystyle{ieee}
\bibliography{trailer_analysis}
}

\end{document}